\documentclass[11pt]{article}
\usepackage[preprint]{acl}
\usepackage{times}
\usepackage{latexsym}
\usepackage[T1]{fontenc}
\usepackage[utf8]{inputenc}
\usepackage{microtype}
\usepackage{inconsolata}
\usepackage{graphicx}

\usepackage{amssymb}
\usepackage{mathtools}
\usepackage{amsthm}
\usepackage{dsfont}
\usepackage{booktabs}
\usepackage{subcaption}
\usepackage{amsmath}
\usepackage{tabularx}
\usepackage{array}        % Enhanced table column definitions (optional but recommended)
\usepackage{pifont}       % Required for the dingbat symbols
\usepackage{multirow}
\newcommand{\cmark}{\ding{51}}  % Defines \cmark as a check
\newcommand{\xmark}{\ding{55}}

\title{Synthetic Contrastive Reasoning for Multi-Table Q\&A}

\author{Ankit Pratap Singh\thanks{Work carried out during the internship at Thoughtworks.} \\
  \small{Iowa State University} \\
  \small{\texttt{sankit@iastate.edu}} \\\And
  Xin Su \\
 \small{Thoughtworks} \\
  \small{\texttt{xin.su@thoughtworks.com}} \\\And
   Phillip Howard \\
  \small{Thoughtworks} \\
  \small{\texttt{phillip.howard@thoughtworks.com}} \\}

\begin{document}
\maketitle

\begin{abstract}
Multi-table question answering requires models to retrieve relevant evidence, link schemas, and perform compositional reasoning across relational tables. Existing multi-table Q\&A resources typically provide questions and final answers but lack reasoning supervision that explains how answers are derived. To address this gap, we construct a synthetic contrastive reasoning-trace dataset for MMQA by generating validated positive traces and plausible negative traces with heterogeneous LLMs. We then use the resulting preference pairs to fine-tune open-weight LLMs with Contrastive Preference Optimization (CPO). Across Qwen3-14B, Mistral-8B, and Llama-3.1-8B, CPO achieves absolute average improvements over Q\&A supervised fine-tuning ranging from 9.7\%-16.3\%, with gains up to 21 percentage points on MMQA. Ablations show that heterogeneous positive and negative trace generators strengthen the contrastive signal, and automated as well as human evaluations indicate that the generated pairs are largely faithful, coherent, and meaningfully contrastive.
\end{abstract}

\section{Introduction}
\begin{figure*}[t]
\includegraphics[width=\textwidth]{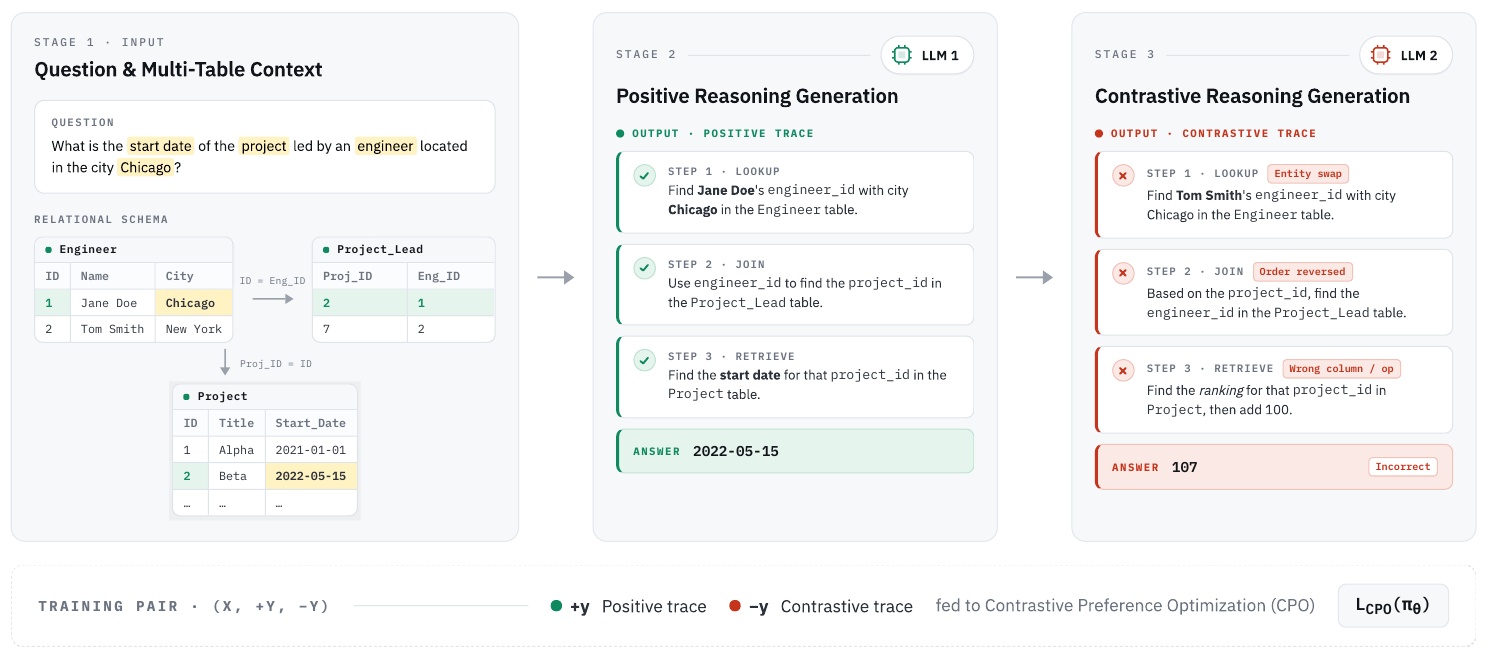}
\caption{Illustration of our approach for generating positive \& contrastive reasoning traces for multi-table Q\&A}
\label{fig:pipeline}
\end{figure*}
Large Language Models (LLMs) are increasingly used in settings where the knowledge needed to answer a question is not contained in the model parameters, but in external structured data. In many enterprise and scientific workflows, this information is stored in relational databases. A useful AI agent must therefore be able to retrieve evidence from tables, identify how schemas relate, and reason over joins, filters, comparisons, and aggregations.
These requirements make multi-table question answering (Q\&A) substantially harder than single-table reasoning. A model may need to identify the relevant tables from a long context, align primary and foreign keys across different schemas, and carry intermediate results through several reasoning hops. As the number of tables grows, contexts become longer, relevant evidence becomes easier to miss \cite{liu2024lost}, and small mistakes in schema linking or constraint checking can lead to incorrect final answers. Although recent work has improved LLMs on single-table reasoning tasks \cite{chainoftable, lei2025reasoning}, multi-table Q\&A remains comparatively underdeveloped.

Reasoning traces are a natural source of supervision for this setting. Prior work on chain-of-thought reasoning shows that step-by-step rationales can improve multi-step problem solving \cite{chainofthought}, and table reasoning often requires exactly this kind of explicit decomposition. However, existing multi-table Q\&A datasets generally provide questions and answers without annotated reasoning paths, and they do not provide paired examples of plausible but incorrect reasoning that can be used for preference optimization.

In this work, we address this gap by producing synthetic contrastive reasoning traces for multi-table Q\&A. We use strong foundation models to generate positive traces that derive the gold answer from the provided tables and negative traces that are fluent and plausible but contain factual or logical errors. We deliberately use heterogeneous LLMs for positive and negative trace generation, which our ablations show creates a stronger contrastive signal than using the same generator for both. We then fine-tune three open-weight LLMs with Contrastive Preference Optimization (CPO), teaching them to prefer correct multi-table reasoning over plausible failure modes.

Our experiments show that this contrastive supervision substantially improves multi-table Q\&A. Across Qwen3-14B, Mistral-8B, and Llama-3.1-8B, CPO achieves absolute average improvements over Q\&A supervised fine-tuning ranging from 9.7\%-16.3\%. On MMQA, the largest absolute gain over Q\&A supervised fine-tuning is 21 percentage points. We also show that CPO outperforms DPO on the same preference pairs, that gains transfer to three-table examples despite training on two-table MMQA data, and that the generated traces are diverse and largely reliable under automated and human evaluation. We will make our code and dataset publicly available to support future research on advancing multi-table reasoning in LLMs. To summarize, our contributions are as follows:
\begin{enumerate}
    \item To the best of our knowledge, we introduce the first contrastive reasoning-trace dataset for multi-table Q\&A, pairing validated positive traces with plausible negative traces generated by heterogeneous LLMs.
    \vspace{-0.2cm}
    \item Through comprehensive experiments, we demonstrate that CPO training on these preference pairs improves multi-table reasoning across three open-weight LLMs and four evaluation datasets.
    \vspace{-0.2cm}
    \item We conduct analyses, ablations, and human evaluations to validate the effectiveness of our synthetic reasoning trace generation approach and the quality of our dataset, validating each LLM-as-judge component against human annotators.
    \vspace{-0.2cm}
    \item We construct a new evaluation dataset for multi-table Q\&A derived from the BIRD benchmark~\citep{li2024can}, introducing a full-table evidence extraction methodology that guarantees answer derivability and automated semantic-consistency verification to filter question--SQL misalignments.
\end{enumerate}
% \vspace{-0.5cm}

\section{Related Work}

\subsection{Table Question Answering}
Table question answering requires models to understand semi-structured data and perform reasoning over it. Early approaches such as TaBERT \cite{tabert} pretrain transformers jointly on textual and tabular data, learning to align table cells with natural language context. Spider \cite{spider} evaluates cross-domain text-to-SQL parsing over multi-table databases, while TableBench \cite{tablebench} focuses on single-table reasoning through direct answer generation.

Recent work extends to multi-table settings. MMQA \cite{mmqa} introduces a multi-table multi-hop QA benchmark that requires models to identify primary and foreign key relationships across tables. MultiTabQA \cite{multitabqa} generates tabular answers by reasoning over multiple related tables. TQA-Bench \cite{tqabench} provides a scalable evaluation framework with context lengths ranging from 8K to 64K tokens. MMTU \cite{mmtu} offers a massive multi-task benchmark for table understanding and reasoning. Despite these advances, most methods rely on implicit reasoning within the model and do not produce explicit multi-step reasoning traces.
% \vspace{-0.4cm}
\subsection{Reasoning with LLMs for Tables}
Chain-of-Thought (CoT) prompting \cite{wei2022chain} elicits step-by-step reasoning from LLMs and improves performance on complex tasks. For tabular data, Chain-of-Table \cite{chainoftable} uses table operations as intermediate reasoning steps, iteratively transforming tables to arrive at answers.

Several works develop table-specialized LLMs. TableLlama \cite{tablellama} fine-tunes open-source models on diverse table tasks to build generalist table understanding capabilities. TableGPT2 \cite{tablegpt2} integrates tabular data into multimodal models. Table-R1 \cite{tabler1} explores inference-time scaling for table reasoning. While these methods improve single-table reasoning, generating synthetic reasoning traces for multi-table QA remains underexplored.

% \vspace{-0.4cm} 
\subsection{Preference Optimization for Reasoning}
Direct Preference Optimization (DPO) \cite{dpo} trains language models to satisfy human preferences without an explicit reward model. However, DPO requires loading both the policy and reference models simultaneously, which increases memory usage and slows training. It also treats all preference pairs equally, ignoring the magnitude of quality differences between responses \cite{amini2024direct, wu2024beta}.

Contrastive Preference Optimization (CPO) \cite{cpo} addresses these limitations by removing the reference model and adding a behavior cloning regularizer. \citet{setlur2406rl} show that training on negative synthetic traces improves math reasoning efficiency for mathematical tasks. Similarly, \citet{chia2023contrastive} show performance gain using both positive and negative traces. Our work extends this idea to multi-table QA, using CPO with contrastive reasoning traces to teach models both correct and incorrect reasoning patterns.

\section{Synthetic Contrastive Reasoning Traces}\label{sec:syn_con_r}

We detail our approach to generating synthetic contrastive reasoning traces for the multi-table Q\&A problem domain. 

\subsection{Problem Setting: Multi-Table Q\&A}
Multi-table Q\&A involves answering a question $[Question]$ given a set of relational tables $[Tables] = [Table_1, Table_2, \ldots, Table_N]$. The task requires identifying and linking relevant information across multiple tables and performing compositional reasoning to generate the correct answer $[Answer]$.

\subsection{Synthetic Positive Reasoning Traces}
To help LLMs solve complex problems involving multiple tables, we generated step-by-step ``reasoning traces'' motivated from Chain-of-Thought (CoT) prompting \cite{wei2022chain, chainoftable} instead of just using simple Q\&A pairs. These traces act as a guide, showing the model exactly how to get from the raw data to the final answer. We primarily used GPT-4o \cite{gpt4o} to create these traces because it is capable of understanding structured data and following logic, but demonstrate how our approach is also effective with open LLMs (see Section~\ref{sec:ablations}). 

\subsection{Prompt Design}
A major challenge in generating synthetic reasoning traces for our task is mitigating hallucinations and ensuring that the model does not rely on outside knowledge when solving the problem; instead, it should only reason over the provided tables to produce the final answer. Therefore, we designed our prompt to encourage the model to focus on the provided input data, instructing it to act as an \textit{``expert at multi-table question answering''}.
Our full prompt (detailed in Table~\ref{tab:prompt_reasoning}) includes three rules to ensure the traces are high-quality:

\textbf{Explicit References}: Every reasoning step must state exactly which table ($TABLE1$ or $TABLE2$) is being used. This forces the model to support its logic with evidence and prevents it from guessing.

\textbf{Step-by-Step Structure}: The output must be formatted as a markdown table (\textbar Step \textbar Output \textbar). This breaks complicated questions down into simple steps, making the process easier for smaller models to learn later.

\textbf{Concise Summaries}: We instructed the model not to copy full rows of data. Instead, it must use short identifiers. This ensures the trace focuses on the logic (e.g., comparing numbers, adding values) rather than just copying text.

\subsection{Validation and Filtering}\label{validation}
We generated candidate reasoning traces for the question and table pairs in our dataset $\mathcal{T}$ using GPT-4o. A trace is defined as ``positive'' ($+y$) if and only if it leads to the correct ground truth answer $[Answer]$.
To verify this, we cannot rely solely on exact string matching, as valid answers often differ in formatting or phrasing (e.g., aliases). Following the evaluation prompt proposed in MMQA \cite{mmqa}, we use LLM as a judge to perform a semantic evaluation. We utilize Gemini 2.0 Flash to compare the model-generated answer against the ground truth.

As detailed in Table~\ref{tab:prompt_eval} of Appendix~\ref{prompt}, the evaluator is prompted to return a score of \textbf{1} if the answers are semantically equivalent and \textbf{0} otherwise. Any trace that receives a score of 0 or fails to adhere to the required markdown formatting is strictly discarded. Thus, our pipeline is specifically designed to mitigate any potential errors by GPT-4o in trace generation.\footnote{58.33\% of generated traces are validated as correct} Rather than relying on the generator being correct, we treat it as a candidate proposer and use the LLM-judge filter to retain only traces whose final answers match the ground truth. We further validate the reliability of this generate-and-filter strategy through human calibration of the judge (Section~\ref{sec:human_eval}), which shows that the filter is conservative (it rejects more correct traces than it lets bad ones through), so the resulting training set is biased toward valid reasoning paths.

\subsection{Synthetic Contrastive Reasoning Traces}
To enable contrastive learning, we require pairs of correct ($+y$) and incorrect ($-y$) reasoning paths. For every validated positive trace generated in the previous step, we synthesized a corresponding negative trace that leads to an incorrect answer.
We used Gemini 2.0 Flash \cite{gemini} to generate these negative traces. Using the prompt template shown in Table~\ref{tab:prompt_contrastive} of Appendix~\ref{prompt}, we provided the model with the correct trace and the original question. The model was instructed to generate a ``contrastive'' trace that is plausible but factually or logically flawed.
Key constraints in the prompt ensure the quality of these negative examples:

\textbf{Active Error Induction}: The model must explicitly alter numbers, logic, or entities to force an incorrect conclusion.

\textbf{Plausibility}: The trace must remain fluent and believable. It should not be nonsensical, as the goal is to teach the student model to detect subtle reasoning errors rather than simple formatting mistakes.

\textbf{Structural Consistency}: The negative trace must follow the same markdown table format as the positive trace to ensure a fair comparison.

\textbf{Justification for Model Selection}: We explicitly chose to use a different model (Gemini 2.0 Flash) for negative traces than the one used for positive traces (GPT-4o). As shown in our ablation study (Figure~\ref{fig:ablation_diversity} of Appendix~\ref{app:additional-results}), using different models for the positive and negative signals consistently yields higher performance than using the same model for both.
Our results suggest that this diversity strengthens the contrastive signal. If the same model generates both traces, the ``style'' of the reasoning remains too similar, making it harder for the student model to distinguish between correct and incorrect logic. By using Gemini 2.0 Flash for the negative traces, we introduce a distinct separation in the reasoning distribution, making the preference optimization process more effective.

\subsection{Training Data}\label{sec:training_data}
We used the MMQA two-table dataset \cite{mmqa} as a basis for generating our synthetic reasoning traces, which contains $2591$ examples of two-table Q\&A pairs. We constructed an evaluation set $\mathcal{E}$ of size $355$ and a training set $\mathcal{T}$ of size $1920$. A few examples were excluded due to context length limitations (128k tokens). The positive and contrastive traces $(+y,-y)$ were generated over the training set $\mathcal{T}$. All models were trained on this dataset $(+y,-y)$.

\section{BIRD Evaluation Set Construction}
\label{sec:bird_construction}
To further evaluate model generalization across diverse domains and data sources, we construct an additional evaluation set derived from the BIRD benchmark~\citep{li2024can}, a large-scale text-to-SQL dataset containing 12,751 question-SQL pairs across 95 databases spanning 37 professional domains. Unlike MMQA (which originates from Spider with Wikipedia-based tables), BIRD uses real-world databases sourced from Kaggle with authentic schema complexity and diverse domain coverage, serving as a challenging out-of-distribution testbed for multi-table reasoning.
We employed three filtering stages to ensure data 
quality, introducing two key contributions: 
(1)~\emph{full-table evidence extraction} to guarantee 
answer derivability, and (2)~\emph{automated semantic 
consistency verification} via LLM-as-judge to filter 
question-SQL misalignments. Table~\ref{tab:bird_pipeline}(Appendix~\ref{app:bird_construction}) summarizes the filtering pipeline and sample counts at each stage.
\paragraph{Stage 1: Multi-Table Filtering.}
From the BIRD training split (9,428 samples), we select queries involving exactly 2 or 3 tables based on the number of JOIN clauses in each SQL query. This yields 6,826 candidates (5,300 two-table and 1,526 three-table), aligning with the multi-hop reasoning setting evaluated on MMQA.

\paragraph{Stage 2: Evidence Extraction with Size Constraints.}
A key challenge in table QA is ensuring that provided evidence is complete: truncated tables may exclude rows necessary to derive the answer, creating unanswerable samples. Unlike prior work that uses fixed row limits (e.g., top-$k$ rows), we extract \emph{full table contents} for each sample, applying size thresholds only to exclude prohibitively large tables (maximum 5k rows per table and 120k total cells per sample). This ensures that any sample in our dataset is answerable from the provided table evidence alone. We execute each gold SQL query on the source database and exclude samples that return empty results. After this stage, 2,258 samples remain with complete, self-contained table evidence.

\paragraph{Stage 3: Semantic Consistency Verification.}
Text-to-SQL datasets may contain misalignments between natural language questions and their SQL annotations~\citep{li2024can}. We employ GPT-5 as a semantic judge to verify that each SQL query exactly captures the question's intent. The judge receives the question, SQL, optional evidence hints from BIRD, and a schema summary (column names and key constraints), outputting ``Yes'' only when confident of semantic equivalence. The full prompt template is provided in Appendix~\ref{app:bird_prompt}. Of 2,258 samples, 1,024 (45.4\%) pass this verification, and these form our final BIRD evaluation set.

\section{Experiments}
We performed a comparative analysis of three fine-tuning strategies across four datasets and three foundational models (Meta-Llama-3.1-8B, Qwen3-14B, and Mistral-8B). The strategies evaluated were: (1) Q\&A SFT (standard Question-Answer pairs), (2) Trace SFT (using positive reasoning traces), (3) Trace DPO (utilizing contrastive reasoning preferences), and (4) Trace CPO (utilizing contrastive reasoning preferences). All experiments were conducted on a single NVIDIA H100 GPU. The evaluation method is the same as discussed in Section~\ref{validation}. Refer to Appendix~\ref{sec:appendix_details} for comprehensive implementation details.
\begin{table*}[t!]
    \centering
    \resizebox{0.7\textwidth}{!}{
    \begin{tabular}{llccccc}
    \toprule
    \textbf{Model} & \textbf{Method} & \textbf{MMQA} & \textbf{MMTU} & \textbf{TableBench} & \textbf{BIRD} & \textbf{Avg.} \\
    \midrule
    \multirow{4}{*}{Qwen3-14B-Base}
      & Q\&A SFT          & 29.0 & 29.0 & 20.0 & 22.0 & 25.0 \\
      & Trace SFT         & 32.0 & 44.0 & 23.0 & 23.0 & 30.5 \\
      & DPO\textsuperscript{$\dagger$}              & 21.0 & 24.2 & 14.0 & 7.8 & 16.8 \\
      & \textbf{CPO (ours)} & \textbf{50.0} & \textbf{56.0} & \textbf{30.0} & \textbf{29.0} & \textbf{41.3} \\
    \midrule
    \multirow{4}{*}{Mistral-8B-Instruct}
      & Q\&A SFT          & 30.0 & 30.0 & 17.0 & 17.0 & 23.5 \\
      & Trace SFT         & 32.0 & 33.0 & 18.0 & 12.0 & 23.8 \\
      & DPO\textsuperscript{$\dagger$}              & 19.0 & \textbf{49.8} & \textbf{27.5} & 9.0 &  26.3\\
      & \textbf{CPO (ours)} & \textbf{43.0} &  46.0 & 22.0 & \textbf{26.0} & \textbf{34.3} \\
    \midrule
    \multirow{4}{*}{Llama-3.1-8B-Instruct}
      & Q\&A SFT          & 28.0 & 31.0 & 18.0 & 16.0 & 23.3 \\
      & Trace SFT         & 38.0 & 40.0 & 19.0 & 18.0 & 28.8 \\
      & DPO\textsuperscript{$\dagger$}              & 20.9 & 38.3 & 20.9 & 6.9 & 21.8 \\
      & \textbf{CPO (ours)} & \textbf{42.0} & \textbf{43.0} & \textbf{24.0} & \textbf{23.0} & \textbf{33.0} \\
    \midrule
    \multicolumn{2}{l}{Table-R1} & 33.6 & 51.0 & 30 & 10.3 & 31.2 \\
    \bottomrule
    \end{tabular}
    }
    \caption{Main results across three base models and four evaluation datasets. Accuracy (\%) on each evaluation set. \textbf{CPO (ours)} and DPO use our synthetic contrastive reasoning traces.}
    \label{tab:main_results}
\end{table*}
 
\subsection{Training Approach}
\label{sec:training_approach}

\citet{setlur2406rl} explored negative synthetic data from a supervised fine-tuned model and used direct preference optimization (DPO) \cite{dpo} for a math-reasoning task.
Given a dataset $\mathcal{D}=\{x_i,+y_i,-y_i\}_{i=1}^{N}$, $x$ is the input (for table Q\&A this is the tables and the question), $+y$ is the positive trace, and $-y$ is the negative trace.
DPO trains a policy $\pi_{\theta}$ using the loss
\resizebox{1\linewidth}{!}{\begin{minipage}{\linewidth}
{\begin{align}
&\mathcal{L}(\pi_{\theta};\pi_{ref}) =\notag\\& \mathds{E}_{(x,+y,-y)\sim\mathcal{D}}\left[\log \sigma\left(\beta\log\frac{\pi_{\theta}(+y|x)}{\pi_{ref}(+y|x)} 
- \beta\log\frac{\pi_{\theta}(-y|x)}{\pi_{ref}(-y|x)}\right)\right]
\end{align}
}
\end{minipage}}

Here, $\pi_{ref}$ is a pre-trained reference model. DPO has two issues. It is memory and speed inefficient because both the policy and the reference model must be loaded at the same time, and each step requires running both models. DPO also fails to account for how different the two responses can be. In some cases the preferred response ($+y$) is only slightly better than the dispreferred one ($-y$). In other cases the dispreferred response is harmful or misleading. DPO still treats all pairs the same \cite{amini2024direct, furuta2024geometric, wu2024beta}, so the model does not learn the real preference gap and can drift away from the true preferred data distribution.

For multi-table Q\&A, these issues are even more important. As shown by \citet{liu2025uncovering}, DPO often fails to improve performance on structured reasoning tasks and can even degrade it. The reasoning chains are long, and many negative traces are only slightly worse than the positive ones. If the negative traces are not contrastive enough, DPO may not learn the right signal, and the model may lose accuracy on multi-step reasoning.

CPO \cite{cpo} helps with both problems. It removes memory and speed issues by setting $\pi_{ref}$ to a uniform prior $U$, so only one model is needed. It also adds a behavior cloning (BC) regularizer \cite{hejna2023contrastive} to keep $\pi_{\theta}$ close to the preferred data. The CPO loss is
\resizebox{1\linewidth}{!}{\begin{minipage}{\linewidth}
\begin{align}
&\mathcal{L}_{CPO}(\pi_{\theta})=\notag\\&-\mathds{E}_{(x,+y,-y)\sim\mathcal{D}}\left[\log\sigma\left(\beta\log\pi_{\theta}(+y|x)-\beta\log\pi_{\theta}(-y|x)\right)\right]\notag \\&-\mathds{E}_{(x,+y)\sim\mathcal{D}}\left[\log(\pi_{\theta}(+y|x))\right]
\end{align}
\end{minipage}}
\vspace{0.5mm}

We use CPO for multi-table Q\&A because it is more stable, faster to train, uses less memory, and keeps the model close to the correct reasoning traces.

\subsection{Main Results}
\label{sec:main_results}

Table~\ref{tab:main_results} provides our main experimental results across all three base models (Qwen3-14B, Mistral-8B, Llama-3.1-8B) and four evaluation datasets. The table also includes DPO-trained models (which utilized the same $(+y,-y)$ pairs as CPO) and Table-R1 \citep{yang2025table}, a recent tabular foundation model, as baselines. All three base LLMs achieve similar levels of performance on the withheld test sets when they are fine-tuned only on the original Q\&A pairs from the MMQA training set. Training on positive reasoning traces generated by GPT-4o improves performance over using Q\&A pairs alone in nearly all cases, demonstrating how explicit reasoning supervision is beneficial. Interestingly, we sometimes observe the largest improvements in out-of-domain performance after training on positive reasoning traces. For example, Qwen3-14B-Base achieves a 3\% improvement on the in-domain MMQA test set and a 15\% absolute improvement in performance on the out-of-domain MMTU evaluation set. In-domain and out-of-domain performance improvements are generally similar for Mistral-8B-Instruct and Llama-3.1-8B-Instruct after training on positive reasoning traces, ranging from 2-3\% improvement for the former and 9-10\% for the latter.
% \vspace{-0.5cm}

Compared to positive trace SFT, DPO\footnote{DPO$^\dagger$ uses the same configuration and SFT checkpoint as CPO.} produces more mixed results. Qwen3-14B performance drops relative to Q\&A-only SFT across all datasets, while the Llama-3.1-8B also sees a drop in performance on MMQA and BIRD. The best results for DPO training were observed for Mistral-8B, where DPO produced large improvements over Q\&A SFT and Trace SFT on MMTU and TableBench, but hurt performance on MMQA and BIRD. As shown by \citet{liu2025uncovering}, DPO often fails to improve performance on structured reasoning tasks and can even degrade it; CPO removes the reference model and adds a behavior cloning regularizer, stabilizing training on long reasoning traces. 

In contrast, CPO training with both positive and negative reasoning traces consistently improves performance across all models and datasets while providing the largest average performance gains.
Compared to training on positive traces only, contrastive training consistently yields higher accuracy across all datasets and models. CPO training also outperforms DPO in all evaluations except MMTU and TableBench for Mistral-8B, without exhibiting the inconsistent performance across datasets of DPO. 
Notably, the performance improvement relative to training only on Q\&A pairs is often much larger for CPO than training only on positive reasoning traces. This is particularly evident in the results for Qwen3 and Mistral, where CPO achieves 13-21\% absolute improvements on the MMQA evaluation set, while training on only positive reasoning traces achieves 2-3\% improvement.
These results demonstrate how exposing the model to incorrect or suboptimal reasoning traces via CPO provides a stronger learning signal than using positive traces alone.

Finally, relative to Table-R1 (a table reasoning model trained with inference-time scaling), our best trained model (Qwen3-14B) performs better across all datasets and achieves an absolute average improvement of 10.1\%. Both Mistral-8B and Llama-3.1-8B also outperform Table-R1 on average across the benchmark suite, with Table-R1 achieving higher performance on MMTU and TableBench but lower accuracy on MMQA and BIRD. This illustrates that even inference-time scaling approaches struggle with multi-table, multi-hop reasoning compared to our CPO-trained models.

\paragraph{Generalization beyond two tables.}
Our training data is drawn from the two-table subset of MMQA (Section~\ref{sec:training_data}), but the approach itself is not restricted to $N=2$. To assess generalization, we evaluated all three base models on the three-table subsets of MMQA and our BIRD evaluation set. Table~\ref{tab:three_table} reports these results. CPO outperforms both Q\&A SFT and Trace SFT on every model and on both datasets, indicating that reasoning patterns learned through CPO on two-table data transfer to harder three-table queries. Absolute accuracies are lower than on the corresponding two-table subsets, which is expected as both the input context length and the number of reasoning hops increase with more tables.

\begin{table}[htbp!]
    \centering
    \resizebox{\columnwidth}{!}{
    \begin{tabular}{llccc}
    \toprule
    \textbf{Dataset} & \textbf{Model} & \textbf{Q\&A SFT} & \textbf{Trace SFT} & \textbf{CPO} \\
    \midrule
    \multirow{3}{*}{MMQA (3-table)}
      & Qwen3-14B     & 28.53 & 27.00 & \textbf{40.88} \\
      & Mistral-8B    & 21.47 & 24.85 & \textbf{34.32} \\
      & Llama-3.1-8B  & 18.97 & 28.84 & \textbf{35.19} \\
    \midrule
    \multirow{3}{*}{BIRD (3-table)}
      & Qwen3-14B     & 50.00 & 50.00 & \textbf{56.25} \\
      & Mistral-8B    & 31.25 & 18.75 & \textbf{43.75} \\
      & Llama-3.1-8B  & 31.25 & 43.75 & \textbf{56.25} \\
    \bottomrule
    \end{tabular}
    }
    \caption{Accuracy (\%) on three-table subsets of MMQA and our BIRD evaluation set. All models are trained on the two-table MMQA subset only; results show out-of-distribution generalization to a higher number of tables.}
    \label{tab:three_table}
\end{table} 
\vspace{-0.4cm}

\section{Analysis}\label{sec:dataset_analysis}
We analyze the quality of our synthetic contrastive reasoning traces through ablation studies (Section~\ref{sec:ablations}), automated evaluation (Section~\ref{sec:auto_eval}), and human calibration of LLM judges (Section~\ref{sec:human_eval}). Diversity metrics and a qualitative example are provided in Appendices~\ref{app:diversity} and~\ref{sec:qualitative}, respectively.
\subsection{Ablations}
\label{sec:ablations}
\paragraph{Impact of using different LLMs for positive and negative traces.}
We study the effect of using different LLMs to generate positive and negative reasoning traces.
From Figure \ref{fig:ablation_diversity} (Appendix~\ref{app:additional-results}), we observe that using different models for positive and negative trace generation consistently leads to higher performance than using the same model for both.
In particular, generating positive traces with GPT-4o and negative traces with Gemini 2.0 achieves the best results across all evaluated base models.
This suggests that increased diversity between positive and negative traces strengthens the contrastive signal, making preference optimization more effective.
Using the same model for both traces results in weaker gains, likely due to reduced separation between correct and incorrect reasoning distributions.
% \vspace{-0.5cm}
\paragraph{Can open-source models produce reasoning traces that are as good as proprietary models?}
We also evaluate the quality of reasoning traces generated by an open-source model.
As shown in Figure~\ref{fig:ablation_quality} in Appendix~\ref{app:additional-results}, traces generated using Qwen-3 30B lead to competitive performance across all base models.
This demonstrates that open-source models are capable of producing useful reasoning traces for contrastive fine-tuning.
However, the highest performance is consistently achieved when using proprietary models, particularly GPT-4o for positive traces and Gemini 2.0 for negative traces.
These results indicate that while open-source models perform well, proprietary models still provide stronger supervision for reasoning trace generation.
\paragraph{Effect of Training Data Size on CPO Performance.}
To analyze the impact of training data size, we train Qwen3-14B-Base with CPO on increasing subsets of our synthetic dataset. Performance consistently improves with more training 
data: 50.0\% with full data vs.\ 31.5\% and 28.2\% 
with 75\% and 50\% of the data, respectively.

\subsection{Automated evaluation of dataset quality}
\label{sec:auto_eval}

We assess trace quality using GPT-5-mini as an 
LLM-as-judge, evaluating each contrastive pair 
along multiple dimensions.
\paragraph{Evaluation Criteria.}
The judge assesses each trace on four dimensions: (1) \textbf{correctness} (0-5): whether the final answer is correct given the tables; (2) \textbf{faithfulness} (0-5): whether claims are supported by table evidence; (3) \textbf{coherence} (0-5): whether the reasoning is clear and logically consistent; and (4) \textbf{hallucination} (boolean): whether any step contains unsupported claims. Additionally, the judge rates the overall \textbf{contrastiveness} (0-5) of each pair, measuring whether the rejected trace serves as an effective negative example for preference learning.

\begin{table}[tb!]
\centering
\resizebox{1\columnwidth}{!}{
\begin{tabular}{lcc}
\toprule
\textbf{Metric} & \textbf{Chosen} & \textbf{Rejected} \\
\midrule
Correctness (mean) & 4.71 & 0.92 \\
Correctness $\geq 4$ (\%) & 92.2 & 3.8 \\
Faithfulness (mean) & 4.80 & 2.74 \\
Faithfulness $\geq 4$ (\%) & 96.6 & 25.9 \\
Coherence (mean) & 4.68 & 3.40 \\
Coherence $\geq 4$ (\%) & 97.8 & 49.3 \\
Hallucination (\%) & 3.1 & 46.1 \\
\midrule
\multicolumn{3}{l}{\textit{Pairwise Judgments}} \\
Chosen wins (\%) & \multicolumn{2}{c}{97.2} \\
Contrastiveness (mean) & \multicolumn{2}{c}{4.03} \\
Contrastiveness $\geq 3$ (\%) & \multicolumn{2}{c}{99.8} \\
\bottomrule
\end{tabular}
}
\caption{LLM-as-judge evaluation results. Correctness, faithfulness, and coherence are scored 0-5 (higher is better). Hallucination is reported as the percentage of traces containing unsupported claims (lower is better).}
\label{tab:judge}
\end{table}
%\vspace{-0.1cm}
\paragraph{Results.}
Table~\ref{tab:judge} shows that chosen traces achieve 
high correctness (4.71/5) and minimal hallucination 
(3.1\%), while rejected traces show substantially lower 
correctness (0.92) and higher hallucination (46.1\%). 
The judge prefers chosen traces in 97.2\% of pairs, 
with a mean contrastiveness score of 4.03, confirming 
our pipeline produces well-separated and plausible 
contrastive pairs.

\subsection{Human Calibration of LLM Judges}
\label{sec:human_eval}
Our pipeline uses LLM-as-judge components at three stages: the Gemini 2.0 Flash answer evaluator (Section~\ref{validation}), the GPT-5 semantic consistency judge for BIRD evaluation set construction (Section~\ref{sec:bird_construction}), and the GPT-5-mini trace quality judge (Section~\ref{sec:auto_eval}). To validate these components, we conducted three human studies.

\paragraph{Gemini answer evaluator.}
We audited $57$ randomly sampled $(\text{generated answer}, \text{gold answer})$ pairs labeled by the Gemini answer evaluator and compared its judgments to human annotations. The evaluator achieves $96.67\%$ precision on the negative class and $70.37\%$ precision on the positive class. The error distribution is asymmetric: 8 false negatives versus 1 false positive. In other words, the judge is conservative---it tends to discard correct answers rather than accept incorrect ones. For the construction of our training set, this skew is favorable: the dominant failure mode is rejecting valid traces (reducing dataset size), not admitting bad traces (contaminating the positive class). As a result, the composition of the retained positive training set is biased toward valid reasoning paths, and our reported performance is likely a lower bound.

\paragraph{GPT-5 semantic consistency judge.}
For the BIRD evaluation set construction, we audited $30$ samples ($15$ accepted, $15$ rejected) by the GPT-5 semantic consistency judge against human annotations. Human and judge labels agree on $70\%$ of samples. Here too the judge is conservative: it rejects $7$ samples that humans would accept, while only $2$ samples accepted by the judge were judged by humans as actually misaligned. Since this judge is used as a quality filter to construct the evaluation set, the conservative behavior means the resulting BIRD evaluation set is a high-precision subset of valid question-SQL pairs at the cost of some recall.

\paragraph{Trace quality judge.}
To validate the trace quality judge used in Section~\ref{sec:auto_eval}, we annotated $30$ pairs of contrastive traces along the same correctness, faithfulness, and coherence dimensions as the LLM judge. Table~\ref{tab:human_eval} compares the resulting mean scores. Human and LLM mean scores agree closely on all three dimensions for both positive and negative traces, supporting the use of the LLM judge in our larger-scale dataset evaluation in Section~\ref{sec:auto_eval}.

\begin{table}[tbp!]
    \centering
    \resizebox{1\columnwidth}{!}{
    \begin{tabular}{lcccc}
    \toprule
    & \multicolumn{2}{c}{\textbf{Positive ($+y$)}} & \multicolumn{2}{c}{\textbf{Negative ($-y$)}} \\
    \cmidrule(lr){2-3} \cmidrule(lr){4-5}
    \textbf{Metric} & \textbf{Human} & \textbf{LLM} & \textbf{Human} & \textbf{LLM} \\
    \midrule
    Correctness   & 5.00 & 4.71 & 1.10 & 0.92 \\
    Faithfulness  & 4.93 & 4.80 & 3.35 & 2.74 \\
    Coherence     & 4.97 & 4.68 & 3.90 & 3.40 \\
    \bottomrule
    \end{tabular}
    }
    \caption{Human vs.\ LLM-judge scores on a sample of $30$ contrastive trace pairs. Scores are 0--5; higher is better.}
    \label{tab:human_eval}
\end{table}

\section{Conclusion}

We showed that synthetic contrastive reasoning traces provide an effective supervision signal for multi-table Q\&A. Across three open-weight LLMs and four evaluation datasets, CPO training on paired positive and negative traces consistently outperformed both Q\&A-only fine-tuning and positive-trace SFT. These gains indicate that models benefit not only from seeing correct multi-table reasoning paths, but also from learning to reject plausible reasoning errors such as missed constraints, incorrect joins, and unsupported aggregations.
We conducted analyses which further show that heterogeneous trace generators improve the contrastive signal, that open-weight generators can produce useful traces, and that the resulting dataset is diverse and largely faithful under automated and human evaluation. Our work highlights how generating synthetic reasoning traces is a promising direction for improving LLM performance on complex reasoning tasks such as multi-table Q\&A.

\section*{Limitations}\label{limitation}

Our highest-quality contrastive traces are produced using GPT-4o for positive traces and Gemini 2.0 Flash for negative traces. Both are commercial APIs, which imposes cost and reproducibility constraints on dataset construction at larger scales. We mitigate this in Section~\ref{sec:ablations} by showing that an open-weight generator (Qwen-3 30B) yields competitive performance.

Three stages of our pipeline (answer validation during trace filtering, semantic consistency verification during BIRD evaluation set construction, and automated dataset-quality evaluation) use LLM-as-judge. Our human studies (Section~\ref{sec:human_eval}) show that these judges are conservative and agree with human annotators on the majority of samples, but the audits are small ($n=30$--$57$), and any residual judge bias propagates into both training data and evaluation.

Our positive trace filter (Section~\ref{validation}) accepts a candidate trace only if its final answer matches the ground truth. It does not verify the correctness of every intermediate reasoning step, so a trace whose intermediate steps are flawed but whose final answer is coincidentally correct can be retained. Our automated and human dataset-quality evaluations indicate that this failure mode is rare ($\geq 92\%$ of chosen traces score $\geq 4$ on correctness, Table~\ref{tab:judge}).

Our training set contains $1920$ contrastive pairs after filtering. While relatively small in size, the consistent gains across three base models and four evaluation datasets demonstrate that this scale is sufficient to elicit the contrastive learning signal. The effectiveness of our training set shows that our approach is data efficient and does not require synthetic data generation at a cost-prohibitive scale to improve model performance.

\clearpage
\bibliography{example_paper}

%%%%%%%%%%%%%%%%%%%%%%%%%%%%%%%%%%%%%%%%%%%%%%%%%%%%%%%%%%%%%%%%%%%%%%%%%%%%%%%
%%%%%%%%%%%%%%%%%%%%%%%%%%%%%%%%%%%%%%%%%%%%%%%%%%%%%%%%%%%%%%%%%%%%%%%%%%%%%%%
% APPENDIX
%%%%%%%%%%%%%%%%%%%%%%%%%%%%%%%%%%%%%%%%%%%%%%%%%%%%%%%%%%%%%%%%%%%%%%%%%%%%%%%
%%%%%%%%%%%%%%%%%%%%%%%%%%%%%%%%%%%%%%%%%%%%%%%%%%%%%%%%%%%%%%%%%%%%%%%%%%%%%%%
\clearpage
\appendix
\section{Prompt}\label{prompt}
We provide complete details of our prompts in Tables~\ref{tab:prompt_reasoning}, \ref{tab:prompt_eval}, and \ref{tab:prompt_contrastive}.

\section{Additional Results}
\label{app:additional-results}

\begin{figure}
    \centering
    % Top Image
    \includegraphics[trim={0 0 0 2.5cm},clip,width=\linewidth]{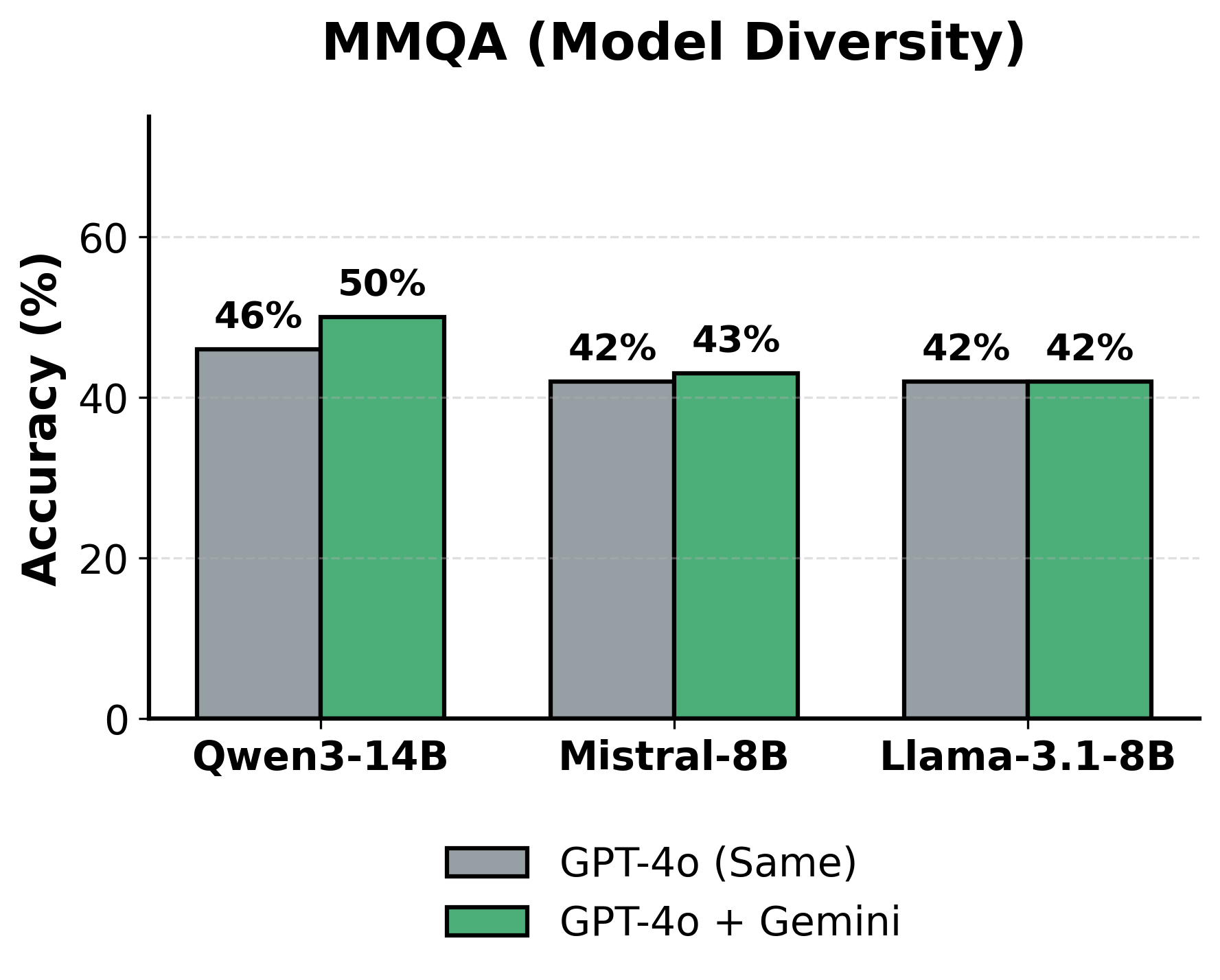}
    %\vspace{0.2cm} % Space between images
    
    % Bottom Image
    \includegraphics[trim={0 0 0 1.6cm},clip,width=\linewidth]{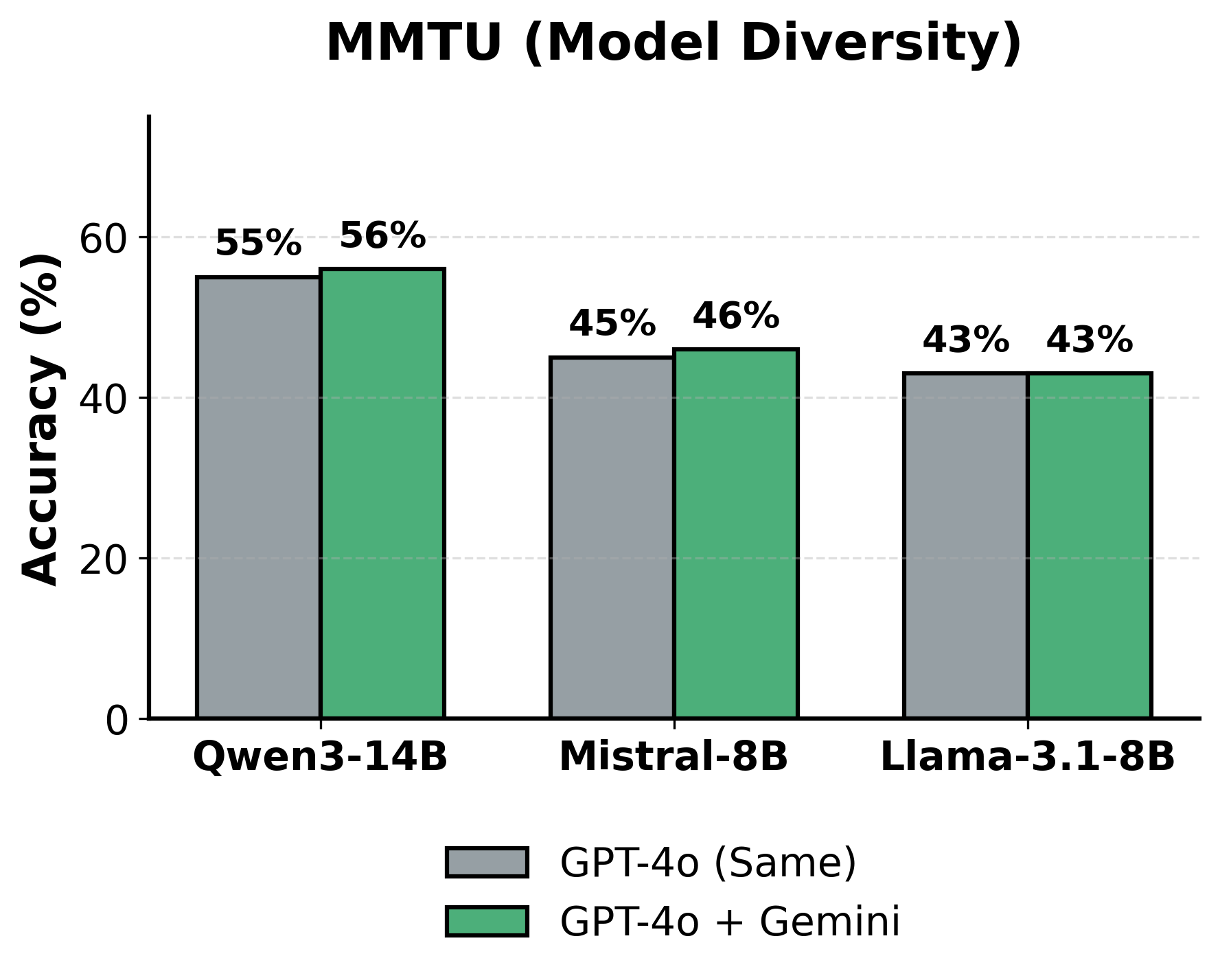}
    
    \caption{\textbf{Impact of Model Diversity.} (Top: MMQA, Bottom: MMTU) We observe that using a distinct model (e.g., Gemini 2.0) for generating reasoning traces consistently outperforms using the same model.}
    \label{fig:ablation_diversity}
\end{figure}

\begin{figure}[t]
    \centering
    % Top Image
    \includegraphics[trim={0 0 0 2.5cm},clip,width=\linewidth]{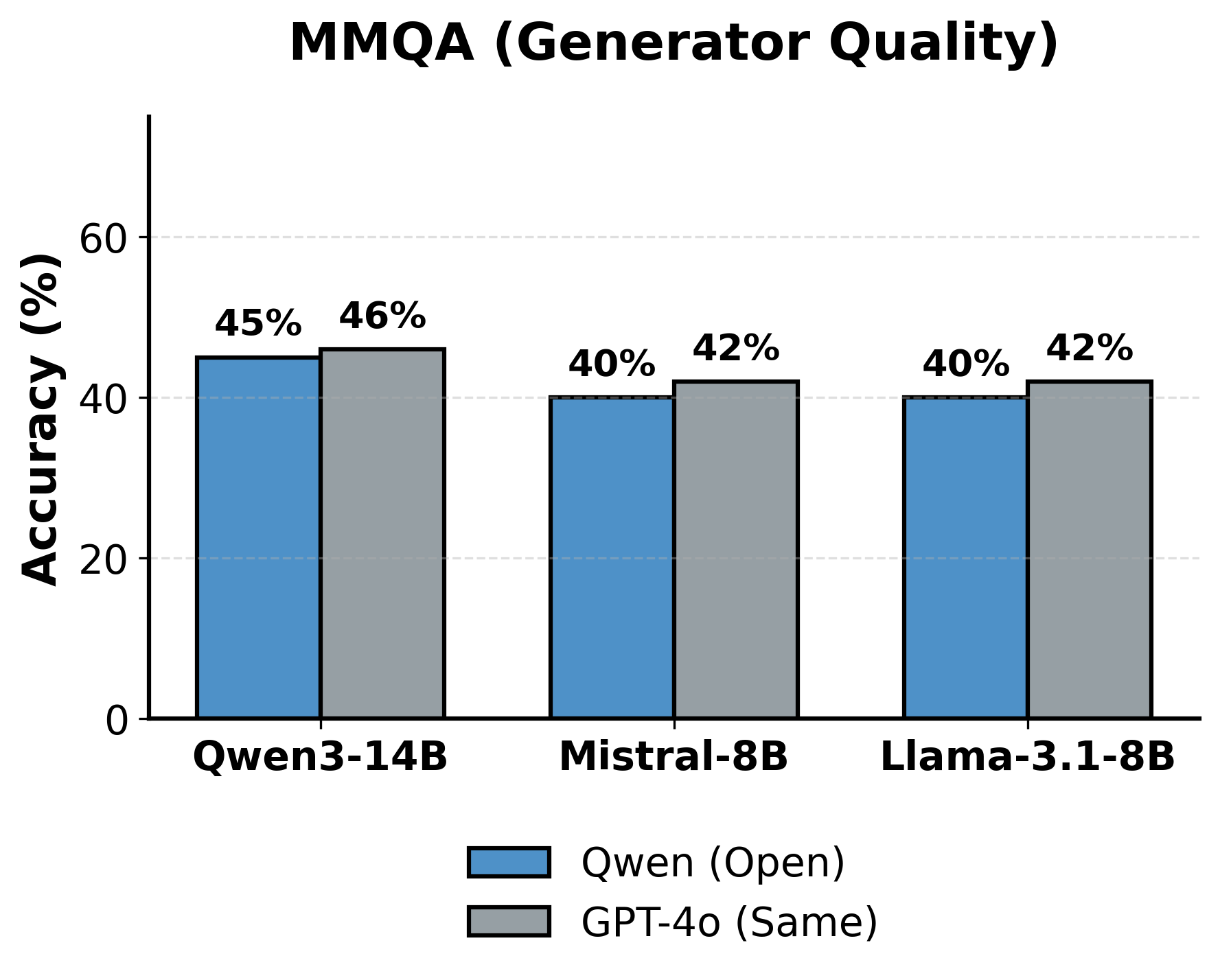}
    \vspace{0.2cm} % Space between images
    
    % Bottom Image
    \includegraphics[trim={0 0 0 2.7cm},clip,width=\linewidth]{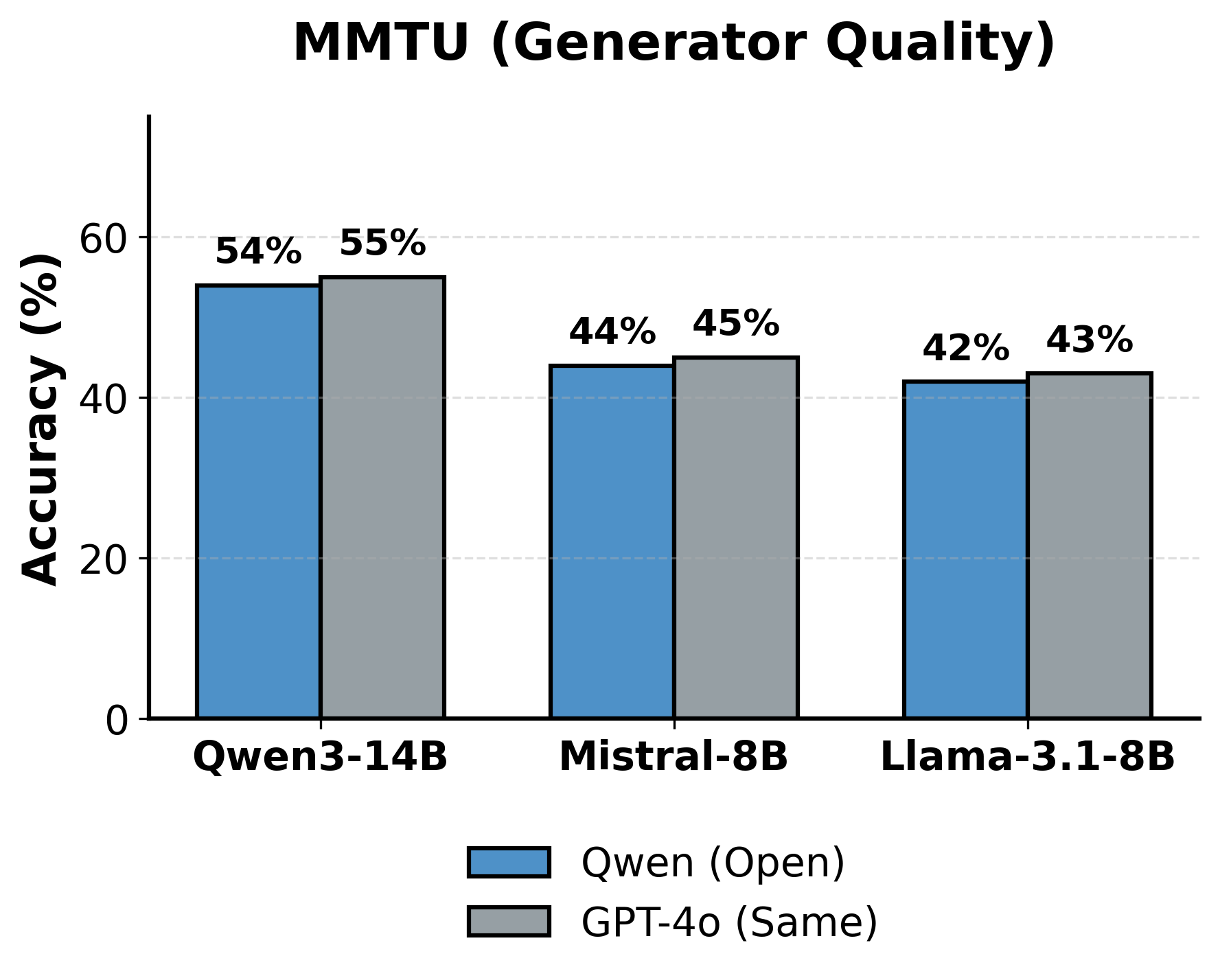}
    
    \caption{\textbf{Generator Quality.} (Top: MMQA, Bottom: MMTU) Performance comparison between open-source and proprietary models. Proprietary model GPT-4o has slightly higher accuracy than open-source Qwen-3 30B. We used the same model for both positive and negative trace generation.}
    \label{fig:ablation_quality}
\end{figure}

Figure~\ref{fig:ablation_diversity} provides an ablation study illustrating the usefulness of using multiple models for generating contrastive reasoning traces. Figure~\ref{fig:ablation_quality} compares the utility of reasoning traces generated by open weights and commercial LLMs.

\section{Experimental Details}\label{sec:appendix_details}
All experiments were conducted on a single NVIDIA H100 80GB GPU.

\paragraph{Models and Training Stages}
We conducted three sets of fine-tuning experiments on three foundational models: Meta-Llama-3.1-8B-Instruct \cite{meta-llama-3.1-8B-Instruct, llama}, Qwen3-14B-Base \cite{qwen}, and Mistral-8B-Instruct \cite{mistral}.

\begin{enumerate}
    \item \textbf{Stage 1 (Q\&A SFT):} The models were fine-tuned using Supervised Fine-Tuning (SFT) on standard Q\&A pairs ($[Question]$, $[Tables]$, $[Answer]$) from the training set $\mathcal{T}$.
    \item \textbf{Stage 2 (Trace SFT):} We fine-tuned the models on the question and synthetically generated positive reasoning traces ($[Question]$, $[Tables]$, $+y$).
    \item \textbf{Stage 3 (CPO Alignment):} We utilized Contrastive Preference Optimization (CPO) \cite{cpo} with contrastive reasoning traces $(+y, -y)$. This stage was initialized using the model weights from Stage 2.
\end{enumerate}

\paragraph{Hyperparameters and Configuration}
We utilized Low-Rank Adaptation (LoRA) \cite{lora} for parameter-efficient fine-tuning across all stages, though with distinct configurations for SFT and CPO to balance plasticity and stability.

\begin{itemize}
    \item \textbf{SFT Configuration (Stage 1 \& 2):} To facilitate the learning of complex reasoning structures, we used a LoRA configuration with $r=1056$ and $\alpha=1056$. We targeted all linear projection layers, including attention (\texttt{q\_proj}, \texttt{k\_proj}, \texttt{v\_proj}, \texttt{o\_proj}) and MLP modules (\texttt{gate\_proj}, \texttt{up\_proj}, \texttt{down\_proj}). Training was performed for 10 epochs with a learning rate of $2\mathrm{e}{-5}$, a cosine learning rate scheduler, 100 warmup steps, and an effective batch size of 8.
    
    \item \textbf{CPO Configuration (Stage 3):} For preference alignment, we adopted a more constrained LoRA configuration with $r=256$ and $\alpha=256$, targeting only the attention modules (\texttt{q\_proj}, \texttt{k\_proj}, \texttt{v\_proj}, \texttt{o\_proj}) while freezing the MLP layers. The CPO training ran for 2 epochs with a learning rate of $2\mathrm{e}{-5}$, a cosine scheduler, 100 warmup steps, and an effective batch size of 8.

    \item \textbf{DPO Configuration:} We train DPO using the same $(+y,-y)$ contrastive pairs and LoRA configuration as CPO, initialized 
from the same SFT checkpoint, with sigmoid loss and $\beta = 0.1$.
\end{itemize}

\paragraph{Evaluation Datasets}
We evaluated on four datasets: MMQA evaluation set $\mathcal{E}$, MMTU \cite{mmtu}, TableBench \cite{tablebench}, and our multi-table QA evaluation set constructed from the BIRD benchmark~\citep{li2024can}. We provide the dataset statistics in Table \ref{tab:dataset_stats}.

\paragraph{License details}
Licenses for datasets and models utilized in our study are as follows: MMQA (CC BY 4.0), MMTU (CC BY 4.0), TableBench (Apache 2.0), BIRD (CC BY-SA 4.0). Models: Qwen3-14B-Base (Apache 2.0), Ministral-8B-Instruct-2410 (Mistral Research License), Llama-3.1-8B-Instruct (Llama 3.1 Community License). All assets are used for research purposes only.

\begin{table*}[h] % Or [t] to place at top of column
    \centering
    % \resizebox{\columnwidth}{!}{ % Scaled to fit single column width
        \begin{tabular}{lcccc}
        \hline
        \textbf{Dataset} & \textbf{Total Tables} & \textbf{Avg Cols} & \textbf{Avg Rows} & \textbf{Total Samples} \\
        \hline
        MMQA & 960 & 5.91 & 522.64 & 355 \\
        MMTU & 1793 & 5.50 & 16.43 & 1793 \\
        BIRD & 2213 & 6.98 & 588.40 & 1024 \\
        TableBench\textsuperscript{*} & 886 & 6.68 & 16.71 & 886 \\
        \hline
        \end{tabular}
    % }
    \caption{Dataset Statistics used in our evaluation.}
    \label{tab:dataset_stats}
    \vspace{2pt}
    \footnotesize
    \raggedright 
    \textit{\textsuperscript{*}TableBench contains single-table queries. For evaluation, the second table input is left empty.}
\end{table*}

\section{BIRD Construction}\label{app:bird_construction}

Table~\ref{tab:bird_pipeline} details the number of samples at each stage of our BIRD evaluation set construction pipeline. 

\begin{table}[t]
\centering
\resizebox{\columnwidth}{!}{
\begin{tabular}{lr}
\toprule
\textbf{Stage} & \textbf{Samples} \\
\midrule
BIRD train (original) & 9,428 \\
Multi-table filtering (2-3 tables) & 6,826 \\
Evidence extraction with size constraints & 2,258 \\
Semantic consistency verification & 1,024 \\
\bottomrule
\end{tabular}
}
\caption{BIRD evaluation set construction pipeline. Each stage progressively filters samples to ensure multi-table structure, complete evidence, and semantic consistency.}
\label{tab:bird_pipeline}
\end{table}

\section{Diversity analysis}\label{app:diversity}
We conduct a comprehensive analysis of our synthetic contrastive reasoning traces to validate their diversity and quality. We examine three dimensions: surface diversity, structural diversity, and pairwise contrastiveness.

\begin{table}[t]
\centering
\resizebox{\columnwidth}{!}{
\begin{tabular}{lcc}
\toprule
\textbf{Metric} & \textbf{Chosen} & \textbf{Negative} \\
\midrule
\multicolumn{3}{l}{\textit{Surface Diversity}} \\
Unique Rate (\%) & 99.7 & 100.0 \\
Vocabulary Size & 5,799 & 8,064 \\
Distinct-1 & 0.025 & 0.031 \\
Distinct-2 & 0.181 & 0.237 \\
Distinct-3 & 0.396 & 0.509 \\
Distinct-4 & 0.590 & 0.736 \\
Median Length (tokens) & 60 & 68 \\
\midrule
\multicolumn{3}{l}{\textit{Structural Diversity}} \\
Median Steps & 5 & 5 \\
Mean Steps & 5.50 & 5.83 \\
Unique Op-Sequence Rate (\%) & 26.9 & 40.5 \\
\bottomrule
\end{tabular}
}
\caption{Surface and structural diversity metrics for chosen (positive) and rejected (negative) reasoning traces.}
\label{tab:diversity}
\end{table}

\begin{table}[t]
\centering
\resizebox{\columnwidth}{!}{
\begin{tabular}{lccccc}
\toprule
\textbf{Metric} & \textbf{Min} & \textbf{P25} & \textbf{Median} & \textbf{P75} & \textbf{Max} \\
\midrule
Token Jaccard & 0.07 & 0.40 & 0.47 & 0.55 & 0.96 \\
Semantic Cosine & 0.41 & 0.80 & 0.85 & 0.90 & 0.99 \\
Step Difference & 0 & 0 & 0 & 1 & 26 \\
\bottomrule
\end{tabular}
}
\caption{Pairwise contrastiveness between chosen and rejected traces. Token Jaccard measures lexical overlap; Semantic Cosine measures embedding similarity using sentence-transformers. P25/P75 denote the 25th/75th percentiles.}
\label{tab:contrastiveness}
% \vspace{0.5em}
\raggedright
%\small{Same-Answer Rate: 4.4\%}
\end{table}
% \vspace{-0.4cm}
\paragraph{Surface Diversity.}
Table~\ref{tab:diversity} presents diversity metrics for our generated traces. Both positive and negative traces exhibit high uniqueness, with 99.7\% and 100\% unique samples, respectively, indicating minimal redundancy in our dataset. Notably, the rejected traces demonstrate greater lexical diversity across all metrics: a larger vocabulary (8,064 vs.\ 5,799 tokens) and consistently higher distinct-n scores. For instance, the distinct-4 score for rejected traces (0.736) substantially exceeds that of chosen traces (0.590). This increased diversity suggests that using different LLMs for positive and negative trace generation---GPT-4o and Gemini 2.0 Flash, respectively---introduces beneficial variation in error patterns and linguistic expressions.
% \vspace{-0.4cm}
\paragraph{Structural Diversity.}
We classify each reasoning step into operation types (e.g., JOIN, FILTER, AGGREGATE, RETRIEVE) based on keyword matching and analyze the resulting operation sequences. As shown in Table~\ref{tab:diversity}, rejected traces exhibit significantly higher structural diversity, with 40.5\% unique operation sequences compared to 26.9\% for chosen traces. Importantly, both trace types maintain similar reasoning depth (median of 5 steps), indicating that negative traces preserve realistic multi-step reasoning structure while introducing diverse error patterns rather than simply truncating or producing shallow responses.
% \vspace{-0.4cm}
\paragraph{Pairwise Contrastiveness.}
To validate that our contrastive pairs provide meaningful training signal for preference optimization, we measure similarity between paired positive and negative traces (Table~\ref{tab:contrastiveness}). The median token-level Jaccard similarity of 0.47 indicates moderate lexical overlap, where pairs share roughly half of their vocabulary while differing in the other half. The median semantic similarity of 0.85 (computed using sentence-transformers) confirms that negative traces remain topically relevant to the original question rather than diverging into unrelated content. The step difference between paired traces has a median of 0, indicating that most negative traces preserve the same reasoning depth as their positive counterparts, with 75\% of pairs differing by at most one step. This balance between similarity and difference is desirable: pairs that are too similar provide weak contrastive signal, while pairs that are too different may confuse the model about what constitutes good reasoning.

\section{BIRD Semantic Consistency Verification Prompt}\label{app:bird_prompt}

Table~\ref{tab:bird_judge_prompt} presents the prompt template used for semantic consistency verification during BIRD evaluation set construction (Section~\ref{app:bird_construction}). The judge receives the natural language question, the gold SQL query, optional evidence hints provided by BIRD, and a schema summary containing table names, column names, and key constraints. The model outputs ``Yes'' only when confident that the SQL query exactly captures the question's intent, and ``No'' otherwise (including cases of uncertainty).

\begin{table*}
    \centering
    \caption{Prompt for generating positive reasoning traces.}
    \label{tab:prompt_reasoning}
    \small
    \begin{tabularx}{\linewidth}{l X}
        \toprule
        \multicolumn{2}{l}{\textbf{System Prompt}} \\
        \midrule
        \multicolumn{2}{p{0.95\linewidth}}{%
            \texttt{You are an expert at multi-table question answering. You are given a [Question] and two tables [TABLE1] and [TABLE2]. Your job is to reason step-by-step and arrive at the correct answer.}
            \par\vspace{2pt}
            \texttt{Each reasoning step must:}
            \vspace{-4pt}
            \begin{itemize}
                \setlength\itemsep{0em} 
                \item \texttt{Refer to either TABLE1 or TABLE2 explicitly}
                \item \texttt{Explain the logical operation concisely}
                \item \texttt{Avoid including full table rows; use short identifiers}
            \end{itemize}
            \vspace{-4pt}
            \texttt{Respond in a markdown table: | Step | Output |, followed by \{"Answer": [...]\}.}
        } \\
        \midrule
        \multicolumn{2}{l}{\textbf{Inputs}} \\
        \midrule
        \textbf{[Question]} & The given multi-hop question. \\
        \textbf{[TABLE1], [TABLE2]} & The input tables (serialized in JSON). \\
        \bottomrule
    \end{tabularx}
\end{table*}

\begin{table*}
    \centering
    \caption{Prompt template for the LLM-as-a-Judge answer evaluation.}
    \label{tab:prompt_eval}
    \small
    \begin{tabularx}{\linewidth}{l X}
        \toprule
        \multicolumn{2}{l}{\textbf{System Prompt}} \\
        \midrule
        \multicolumn{2}{p{0.95\linewidth}}{%
            \texttt{You are an Answer evaluator. Measure the semantic similarity between [Generated Answer] and [Gold Answer]. Return 1 if they mean the same (allowing for aliases like "Messi" == "Lionel Messi"), else return 0.}
        } \\
        \midrule
        \multicolumn{2}{l}{\textbf{Inputs}} \\
        \midrule
        \textbf{[Generated Answer]} & The model-generated answer to be evaluated.\\
        \textbf{[Gold Answer]} & The gold (reference) answer.\\
        \bottomrule
    \end{tabularx}
\end{table*}

\begin{table*}
    \centering
    \caption{Prompt template for generating contrastive (negative) reasoning traces.}
    \label{tab:prompt_contrastive}
    \small
    \begin{tabularx}{\linewidth}{l X}
        \toprule
        \multicolumn{2}{l}{\textbf{System Prompt}} \\
        \midrule
        \multicolumn{2}{p{0.95\linewidth}}{%
            \texttt{You are an expert reasoning trace generator. Your task is to generate a contrastive reasoning trace that is wrong.}
            \par\vspace{2pt}
            \texttt{Requirements:}
            \vspace{-4pt}
            \begin{itemize}
                \setlength\itemsep{0em}
                \item \texttt{Make changes at each step to induce errors (alter numbers, logic, or entities).}
                \item \texttt{The contrastive trace should be plausible but incorrect.}
                \item \texttt{You may reorder steps if it helps make the trace more different.}
                \item \texttt{Follow the markdown table format: | Step | Output | ... followed by \{"Answer": [...]\}}
            \end{itemize}
        } \\
        \midrule
        \multicolumn{2}{l}{\textbf{Inputs}} \\
        \textbf{[Prompt]} & The original multi-table QA prompt. \\
        \textbf{[Correct Trace]} & The correct reasoning trace (positive example). \\
        \bottomrule
    \end{tabularx}
\end{table*}

\begin{table*}[h]
\centering
\caption{Prompt template for BIRD semantic consistency verification using GPT-5 as judge.}
\label{tab:bird_judge_prompt}
\small
\begin{tabularx}{\textwidth}{lX}
\toprule
\multicolumn{2}{l}{\textbf{System Prompt}} \\
\midrule
\multicolumn{2}{p{0.95\textwidth}}{%
\texttt{You are a strict SQL semantics auditor.}
\par\vspace{4pt}
\texttt{Task:}
\par
\texttt{- Given a natural language Question and a gold SQL query, decide whether the SQL's semantics EXACTLY match the Question's intent.}
\par
\texttt{- Evidence is optional; if provided, use it only to disambiguate terms/constraints.}
\par
\texttt{- Schema is optional; if provided, it contains column names (and sometimes key hints). Use it only to interpret the SQL correctly.}
\par\vspace{4pt}
\texttt{Decision rule:}
\par
\texttt{- Answer "Yes" ONLY if you are confident there is NO semantic mismatch.}
\par
\texttt{- Otherwise answer "No" (including any uncertainty).}
\par\vspace{4pt}
\texttt{Treat as mismatch (non-exhaustive):}
\par
\texttt{- Missing/extra filtering conditions or constraints}
\par
\texttt{- Wrong aggregation level/unit (rows vs entities), join multiplicity causing miscounts}
\par
\texttt{- Incorrect join logic, grouping, DISTINCT, LIMIT/OFFSET, ORDER BY direction}
\par
\texttt{- Answering a related but different question}
\par\vspace{4pt}
\texttt{Output format:}
\par
\texttt{- Output exactly one token: Yes or No.}
} \\
\midrule
\multicolumn{2}{l}{\textbf{User Prompt Structure}} \\
\midrule
\textbf{[db\_id]} & The database identifier from BIRD. \\
\textbf{[Question]} & The natural language question from BIRD. \\
\textbf{[Evidence]} & Optional evidence/hints provided by BIRD for disambiguation. \\
\textbf{[SQL]} & The gold SQL query to be verified. \\
\textbf{[Schema]} & Schema summary: table names with column lists, plus primary/foreign key annotations. \\
\bottomrule
\end{tabularx}
\end{table*}

\begin{table*}
    \centering
    \small
    \renewcommand{\arraystretch}{1.3}
    \begin{tabularx}{\linewidth}{X|X}
        \toprule
        \multicolumn{2}{l}{\textbf{Question:} What is the total price for products in \textbf{level 8} with \textbf{attribute 4} and value `1'?} \\
        \multicolumn{2}{l}{\textbf{Gold Answer:} 687.59} \\
        \midrule
        \textbf{Positive Reasoning Trace ($+y$)} & \textbf{Negative Reasoning Trace ($-y$)} \\
        \midrule
        1. Identify entries in \texttt{TABLE2} with attribute\_id 4 and value '1'. Result: \textbf{ID 2, ID 5}. & 1. Identify entries in \texttt{TABLE2} with attribute\_id 4 and value '1'. Result: \textbf{ID 2, ID 5}. \\
        \hline
        2. Look up these IDs in \texttt{TABLE1} to check \texttt{catalog\_level\_number}. & 2. \textbf{Retrieve prices} for these IDs from \texttt{TABLE1}. \textcolor{red}{\textit{(Skipped constraint verification)}} \\
        \hline
        3. \textbf{Verification:} & 3. Price for ID 2 is 687.59. \\
        \quad - ID 2 is Level 8 \textcolor{teal}{(\cmark Match)} & \\
        \quad - ID 5 is Level 9 \textcolor{red}{(\xmark Mismatch - Exclude)} & 4. Price for ID 5 is 616.22. \\
        \hline
        4. Sum prices of matching entries (only ID 2). & 5. \textbf{Sum prices} of both entries. \\
        \hline
        5. Total: 687.59 & 6. Total: $687.59 + 616.22 = 1303.81$ \\
        \midrule
        \textbf{Final Answer:} 687.59 \textcolor{teal}{\cmark} & \textbf{Final Answer:} 1303.81 \textcolor{red}{\xmark} \\
        \bottomrule
    \end{tabularx}
    \caption{Comparison of constraint verification.}
    \label{tab:constraint_failure}
\end{table*}

\section{Qualitative Analysis}
\label{sec:qualitative}

To understand the effectiveness of Contrastive Preference Optimization (CPO), we analyze a specific example from our training data where the model must distinguish between correct and incorrect reasoning. Table \ref{tab:constraint_failure} presents a case involving a multi-table query.

In this example, the negative trace ($-y$) demonstrates a common error where the model fails to verify all constraints before calculating the answer. The question requires the model to filter products based on two conditions: an attribute in \texttt{TABLE2} and a catalog level in \texttt{TABLE1}.

As shown in Table \ref{tab:constraint_failure}, the negative trace correctly performs the first step: it identifies both items (ID 2 and ID 5) that match the attribute criteria. However, it skips the second step. It does not check if these items also belong to ``level 8.''

Instead, the model assumes that because the items matched the first rule, they are valid. It proceeds to sum the prices of both items, including ID 5 (which is actually Level 9). This leads to an incorrect total. Meanwhile, the positive trace explicitly checks the level of each candidate and correctly removes ID 5 before doing the math. By training on these examples, CPO teaches the model to verify every constraint in the question before providing a final answer.
\end{document}